\pdfoutput=1

\documentclass[11pt]{article}

\usepackage{acl}

\usepackage{times}
\usepackage{latexsym}

\usepackage[T1]{fontenc}

\usepackage[utf8]{inputenc}

\usepackage{microtype}

\usepackage{graphicx}

\usepackage{booktabs}
\usepackage{amsmath}
\usepackage{amssymb}   
\usepackage{bbm}

\usepackage{array,multirow,colortbl}
\newcolumntype{C}[1]{>{\centering\arraybackslash}m{#1}}

\usepackage{microtype}

\title{You can't pick your neighbors, or can you? \\ When and how to rely on retrieval in the $k$NN-LM}

\author{Andrew Drozdov\thanks{~~Corresponding author: adrozdov@cs.umass.edu},
  ~Shufan Wang,
  ~Razieh Rahimi,\\
  {\bf ~Andrew McCallum,
  ~Hamed Zamani,
  ~and~Mohit Iyyer}\\
  \AND\\[-4ex]
Manning College of Information and Computer Sciences\\University of Massachusetts Amherst}

\date{}

\begin{document}
\maketitle
\begin{abstract}

Retrieval-enhanced language models (LMs), which condition their predictions on text retrieved from large external datastores, have recently shown significant perplexity improvements compared to standard LMs. One such approach, the $k$NN-LM, interpolates any existing LM's predictions with the output of a $k$-nearest neighbors model and requires no additional training. In this paper, we explore the importance of lexical and semantic matching in the context of items retrieved by $k$NN-LM. 
We find two trends: (1) the presence of large overlapping $n$-grams between the datastore and evaluation set plays an important factor in strong performance, even when the datastore is derived from the training data; and (2) the $k$NN-LM is most beneficial when retrieved items have high semantic similarity with the query.
Based on our analysis, we define a new formulation of the $k$NN-LM that uses retrieval quality to assign the interpolation coefficient. We empirically measure the effectiveness of our approach on two English language modeling datasets, Wikitext-103 and PG-19. Our re-formulation of the $k$NN-LM is beneficial in both cases, and leads to nearly 4\% improvement in perplexity on the Wikitext-103 test set.

\end{abstract}

\section{Introduction}

Recently, a new class of language models (LMs) that are augmented with \emph{retrieval} capabilities have led to substantial improvements over standard neural LMs \cite[inter alia]{Lewis2020RetrievalAugmentedGF,He2020LearningSP,Yogatama2021AdaptiveSL,Borgeaud2021ImprovingLM,Transformers2022MET,Thoppilan2022LaMDALM}.
Furthermore, LMs with retrieval warrant investigation as they provide benefits for many tasks \cite{Zamani2022REML}.
These approaches generally involve a backbone neural LM that interacts with a retrieval component of varying complexity to find relevant documents. In this work, we analyze and improve a specific and simple type of retrieval-enhanced language model, the $k$NN-LM originally proposed by~\citet{khandelwal20generalization}.

The $k$NN-LM is non-parametric --- it works by retrieving instances from an external datastore at each decoding timestep, and it improves language model performance without requiring additional training. In essence, the $k$NN-LM interpolates a base LM's predicted probability distribution of the next word with a distribution formed by \emph{retrieving} vectors similar to the current hidden state.
$k$NN-LM includes two tunable hyperparameters: the number of items to retrieve ($k$) and an interpolation coefficient ($\lambda$). The method's effectiveness depends crucially on source and size of the retrieval datastore: it is most effective when using a very large datastore with orders of magnitude more tokens than seen in the training corpus, but~\citet{khandelwal20generalization} also observe improvements with smaller datastores.

\begin{figure}[!t]
\centering
\includegraphics[width=\linewidth]{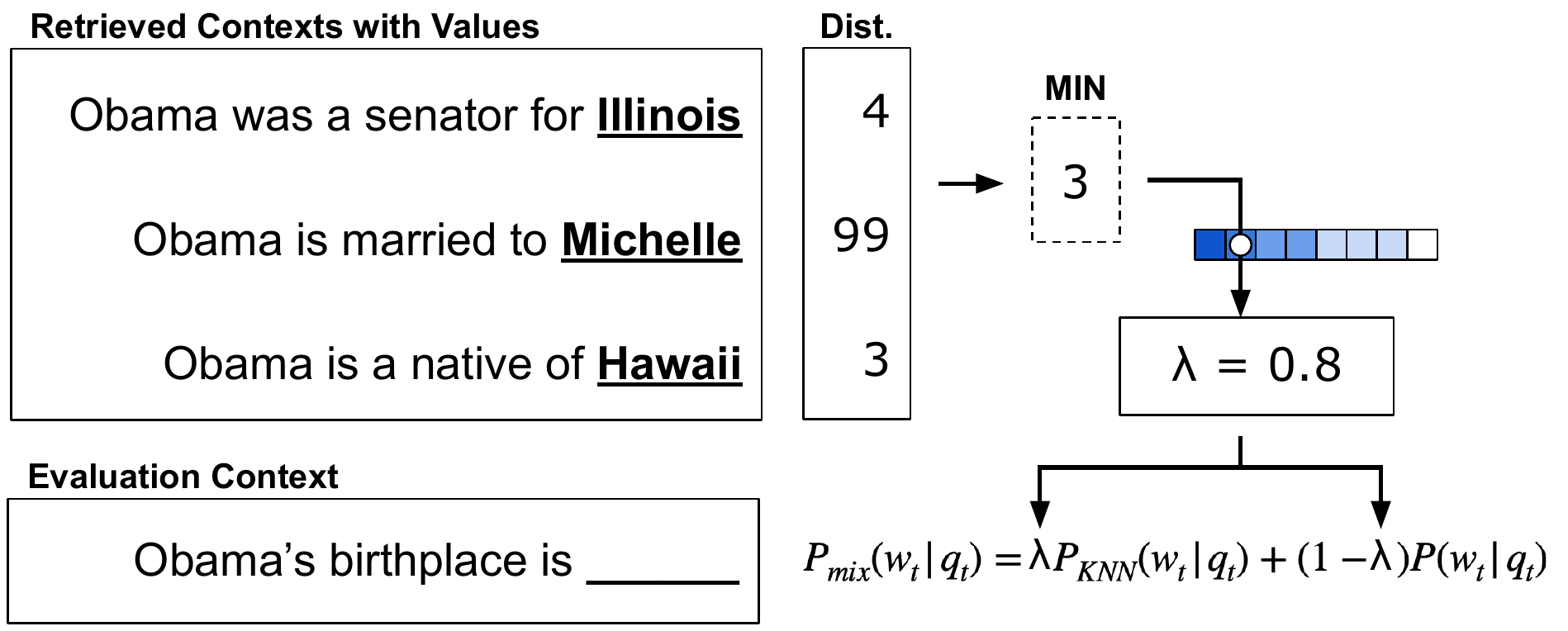}
\caption{We present an extension to $k$NN-LM that conditions the interpolation coefficient ($\lambda$) on the semantic similarity of retrieved contexts.}
\label{fig:adaptive_coeff}
\end{figure}

Modern neural models have massive capacity to memorize their training data \cite{Zhang2017UnderstandingDL}. Nonetheless, simply using an LM's training corpus as the source for the datastore works well for $k$NN-LM, as test perplexity on the Wikitext-103 dataset decreases substantially from 18.65 to 16.12. However, it remains unclear how and why the $k$NN-LM achieves these improvements. Which types of tokens and contexts does it improve most on? As an effort to answer this question and motivate new more effective methods to enhance LMs with retrieval we analyze the $k$NN-LM's behavior with respect to parts of speech, semantic similarity between context and retrievals, and lexical overlap. 

Among others, our analysis reveals the $k$NN-LM is helpful beyond factual knowledge (i.e. proper nouns), and improves perplexity across many word types, so it would be difficult to extend $k$NN-LM using syntactic information alone. On the other hand, we find the performance of the $k$NN-LM highly correlates with lexical similarity between the context and retrieved items, although this is somewhat domain specific and does not fully explain its strong performance. Semantic similarity is nearly as accurate a predictor of $k$NN-LM performance as lexical similarity, making it a strong candidate to extend the $k$NN-LM.

Based on our analysis, we devise a simple scheme to extend the $k$NN-LM following the intuition that when retrieval quality is high (measured by semantic similarity), then the model should rely more heavily on the $k$NN-based prediction. Since retrieval in the $k$NN-LM is latent, we use semantic similarity as a proxy to measure retrieval relevance. Concretely, our method is an \textit{adaptive} version of $k$NN-LM that assigns the interpolation coefficient according to \textit{retrieval quality} (see Figure \ref{fig:adaptive_coeff}). While it introduces new hyperparameters,  we show that the additional hyperparameter tuning comes at negligible cost. Importantly, our empirical results demonstrate that our newly introduced re-formulation of $k$NN-LM is beneficial for both encylopedic text and book data, and leads to an improvement of nearly 4\% perplexity over the the vanilla $k$NN-LM, measured on the English language modeling Wikitext-103 test set. Broadly, we hope our insights and methods helps to facilitate future development of retrieval-augmented LMs.

\section{Language Modeling with $k$NN-LM}

The $k$NN-LM improves over a base language model by explicitly \textit{memorizing} the LM's training data. It stores exact sentences from the training data in its datastore that can be accessed during language model inference to produce a $k$-nearest neighbor next word distribution that is interpolated with the base model's prediction. Interpolation is preferred for similar reasons as approximate matrix factorization in collaborative filtering --- the universe of text patterns is sparse and lossless compression of the training data alone is not sufficient to model new patterns. In this section, we explain the specifics of the $k$NN-LM's inner workings in order to guide our analysis.

\subsection{General Approach}
\label{sec:knnlm_approach}

The $k$NN-LM \cite{khandelwal20generalization} is a language model with a retrieval component. Like all language models, it predicts the the word at time step $t$ conditioned on the history of words: $P(w_t | w_{0}, w_{1}, \dots, w_{t-1})$. Neural language models encode the history of words using a vector $h$: $P(w_t | h_{t-1})$. What makes the $k$NN-LM novel is that it uses a pretrained language model to encode a collection of documents, and then retrieves documents from this collection based on vector similarity in order to improve its next word prediction. Notably, the retrieval is completely latent --- no supervised ranking information is used and  documents are  retrieved using semantic similarity.

The $k$NN-LM follows a particular way of encoding the collection of documents into a datastore. Consider document $x_i$ consisting of $n$ words. The $k$NN-LM encodes the first $n-1$ words as a vector and this becomes the \textbf{key} of document $x_i$, referred to as $k_i$. The $n$-th word is saved as the \textbf{value} $v_i$. In practice, and since $k$NN-LM is used for language modeling, a sequence with $n$ words is recorded as $n-1$ documents: for any $t \leq n$, a document whose key is words $w_1$ to $w_{t-1}$ and value is $w_t$ is built.

After the datastore is built, the $k$NN-LM is evaluated on a dataset with $m$ words, predicting words from left-to-right. Retrieval in $k$NN-LM is done by measuring Euclidean distance $d(., .)$ between vector encodings of the \textbf{query} $q_j$ (corresponding to the context of the $j$-th word in the evaluation data) and the keys in the datastore. The values from retrieved documents define a new distribution of the next word:

\begin{align} \label{eq:knn_prob}
    P_{KNN}(w_t | q_t) &\propto \sum_{(k_i, v_i)} \mathbbm{1}_{w_t = v_i} \exp(-d(k_i, q_t))
\end{align}

The best performance typically involves mixing the original and $k$NN-based word distributions using a tunable hyperparameter $\lambda$:

\begin{align*} 
    P^{'}(w_t | q_t) &= \lambda P_{KNN}(w_t | q_t) + (1-\lambda) P(w_t | q_t)
\end{align*}

The $\lambda$ is fixed, yet it would be beneficial if $\lambda$ was conditioned on a per-token basis. We present an approach along these lines in the next section.

\begin{figure}[!t]
\centering
\includegraphics[width=\linewidth]{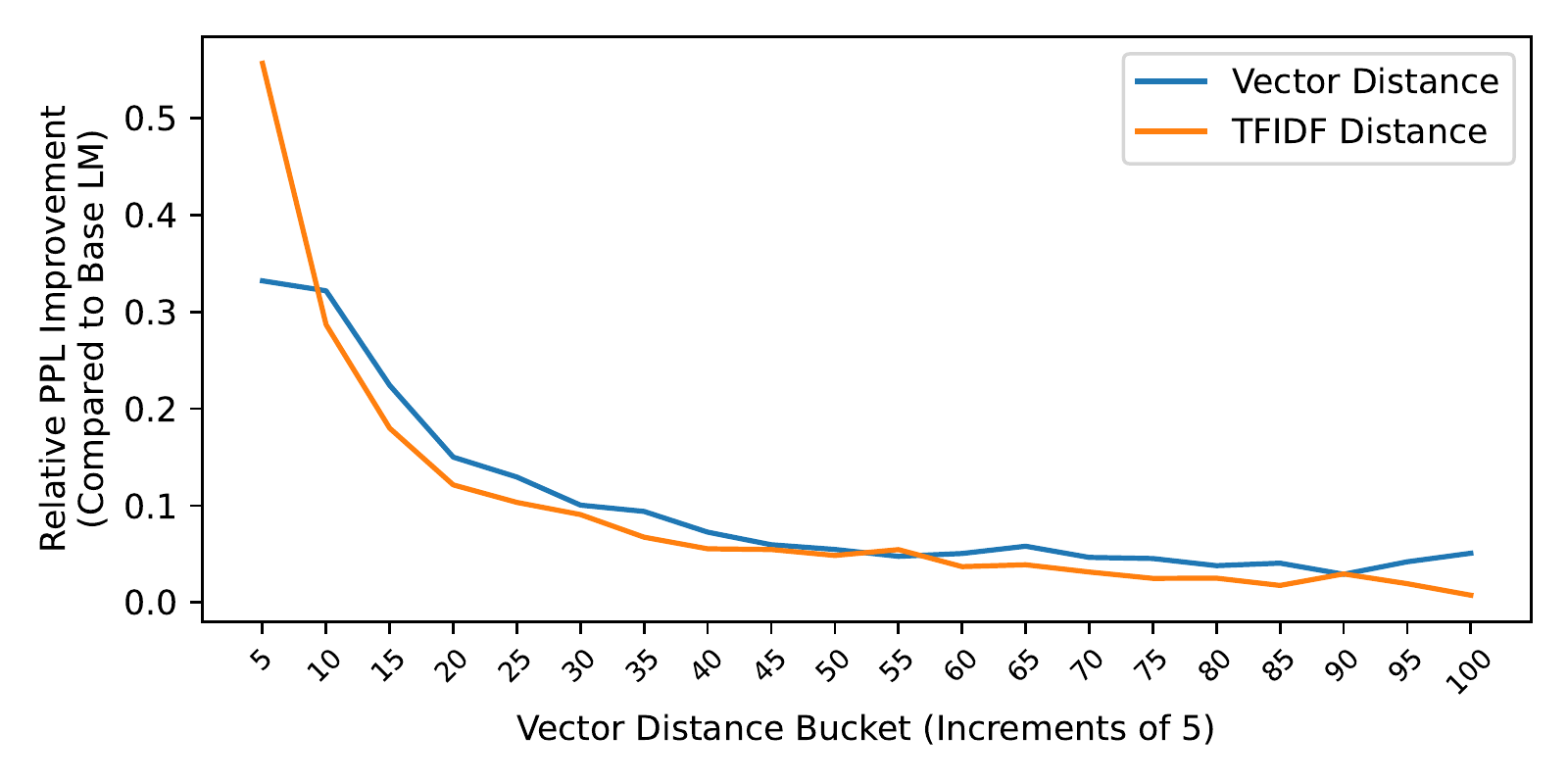}
\caption{Relative perplexity improvement of $k$NN-LM compared to the base language model measured on the Wikitext-103 validation set. Queries are bucketed by semantic similarity of the top retrieved item, which operates as a proxy for retrieval quality.}
\label{fig:analysis_vecdist}
\end{figure}

\section{Analysis: When is $k$NN-LM effective?}

\label{sec:initial_analysis}

In the original $k$NN-LM work, the authors made qualitative observations that the model generally helps for rare patterns, factual knowledge, and names \cite{khandelwal20generalization}. In this section we perform automated analysis to more specifically understand when $k$NN-LM is beneficial, with the aim to uncover systematic behavior that can be leveraged to extend $k$NN-LM and improve its effectiveness at next word prediction.

\subsection{Semantic Similarity of Retrieved Items}

\label{sec:analysis_vecdist}

The $k$NN-LM encodes the context into a fixed-length query vector and uses this to retrieve semantically similar contexts from the datastore. A priori, it's difficult to know when retrieval will be helpful, but perhaps there is a higher chance for usefulness if the result closely matches the query.

Figure \ref{fig:analysis_vecdist} examines this intuition a posteriori on the Wikitext-103 validation set. We bucket queries according to their semantic similarity with their top retrieved item, then report the relative perplexity improvement of the $k$NN-LM over the base model separately for each bucket.\footnote{Bucketing provides a coarse view of $k$NN-LM performance. Language modeling is a near impossible task with many valid continuations for each prediction, so aggregate performance can be more informative.} The queries are sorted by the associated semantic similarity, then divided into 20 equally sized bucket. The first contains the 5\% that have the highest semantic similarity with their top retrieved item. The plot in Figure \ref{fig:analysis_vecdist} clearly indicates that $k$NN-LM is most beneficial in the buckets with high semantic similarity, supporting the hypothesis that semantic similarity is a proxy for retrieval quality.

\begin{table}[t!]
\setlength\tabcolsep{4pt}
\begin{center}
\begin{tabular}{ r | C{10mm} | C{10mm} | C{10mm} | C{10mm} }
\toprule
 & Dev & Dev-8 & Test & Test-8 \\
\midrule
{\small \bf Wikitext} & & & & \\
BaseLM & 17.96 & 17.96 & 18.65 & 18.65 \\
$k$NN-LM & 16.06 & 17.28 & 16.12 & 18.05 \\
Ours & 15.72   & 17.26 & 15.50 & 18.03 \\
\midrule
{\small \bf PG-19} & & & & \\
BaseLM & 60.83 & 60.83 & 50.95 & 50.95 \\
$k$NN-LM & 52.49 & 53.34 & 43.93 & 44.97 \\
Ours   & 52.08 & 53.06 & 43.58 & 44.78 \\
\bottomrule
\end{tabular}
\end{center}
\caption{Perplexity on Wikitext-103 and PG-19 datasets. Dev-8 and Test-8 contain the same data as Dev and Test, but overlapping $n$-grams ($n \geq 8$) with the evaluation data have been removed from the $k$NN-LM datastore. Our method (\S\ref{sec:adaptive_coeff}) uses retrieval quality to interpolate between $k$NN and base LMs.}
\label{tab:overlap_ppl}
\end{table}

\subsection{Lexical Overlap}

\label{sec:analysis_bow}

Another possible proxy for relevance is lexical overlap. Rather than assign queries to buckets using semantic similarity derived from neural network hidden states, we first convert contexts into TFIDF vectors (using 32-token trailing window), which are a popular and effective bag-of-words representation \cite{chen2017reading}. We use the same neighbors as before, but now assign buckets using distance between TFIDF vectors. The relative perplexity for this setting is reported in Figure \ref{fig:analysis_vecdist}, and aligns well with what we saw using semantic similarity in the previous subsection. This suggests that $k$NN-LM is also beneficial when query contexts have high lexical overlap with the datastore contexts.

To further examine the role of lexical matching in the performance of $k$NN-LM, we rebuild the index used for retrieval in a way that minimizes lexical overlap. The keys are identical to before, but we ignore contexts that include large overlapping $n$-grams ($n \geq 8$) with the evaluation data.\footnote{To ensure the $n$-gram context does not leak into the datastore, we follow \citet{gpt} and ignore tokens corresponding to a 200-token window centered around the $n$-gram.} In Table~\ref{tab:overlap_ppl}, we compare the original with this new restricted datastore on Wikitext-103. Even with these lexically similar contexts removed, the $k$NN-LM still provides some benefit (although severely diminished), so lexical similarity alone does not fully explain performance.

\subsection{Part-of-Speech Tags}

\label{sec:syntactic_analysis}

Another lens, syntax, sheds light on $k$NN-LM performance outside of document relevance.  To further understand which types of words benefit most from $k$NN-LM,  we group tokens by their part-of-speech. Then we compute validation perplexity separately for each group using both the base language model and the $k$NN-LM. To get part-of-speech tags, we segment the data into sentences and label words using the tagger from Stanza\footnote{\href{https://stanfordnlp.github.io/stanza/}{https://stanfordnlp.github.io/stanza/}} with the universal dependencies output space. We include categories with frequency greater than 1K in the Wikitext-103 validation data.

\begin{figure}[!t]
\centering
\includegraphics[width=\linewidth]{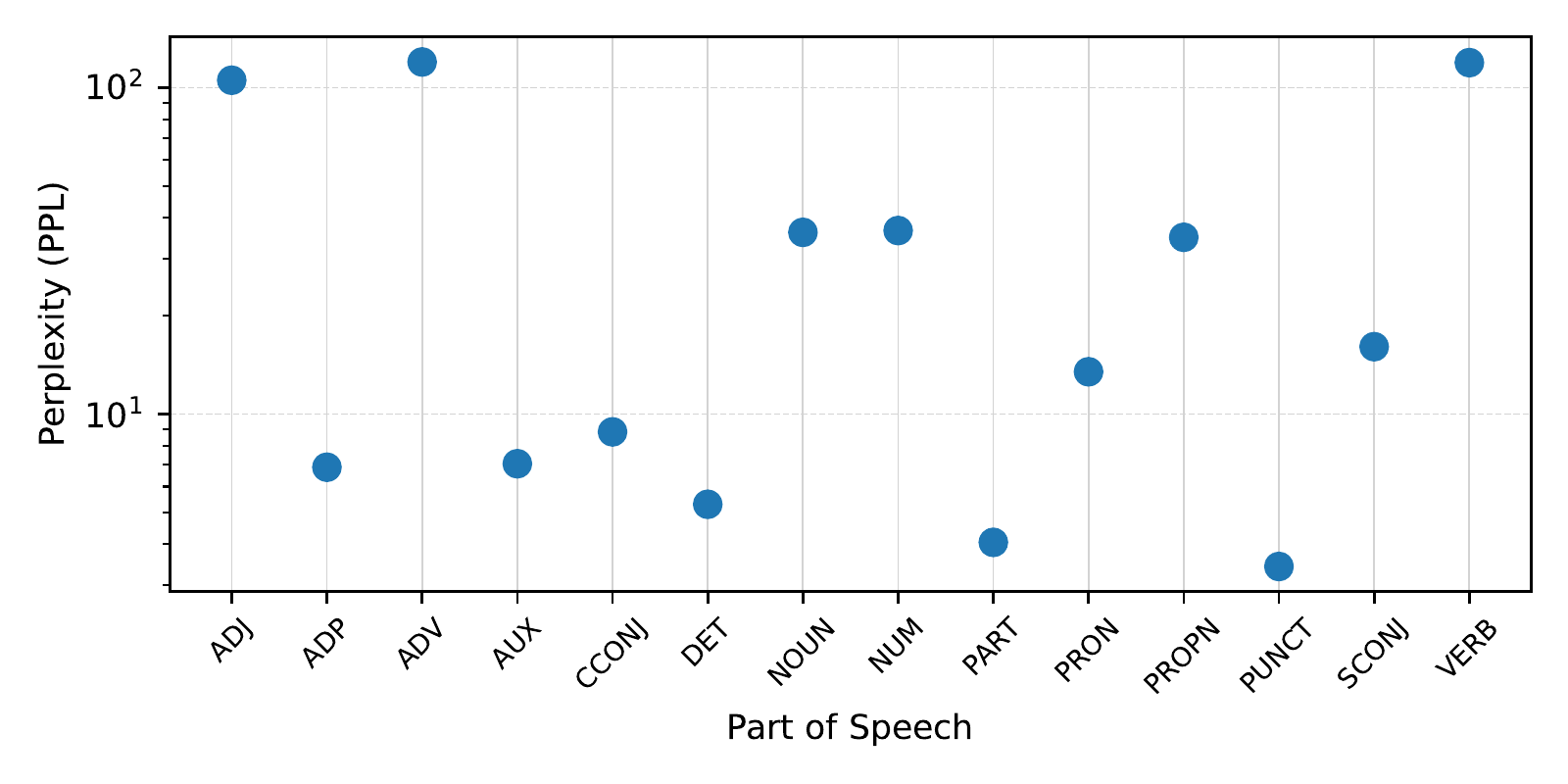}
\includegraphics[width=\linewidth]{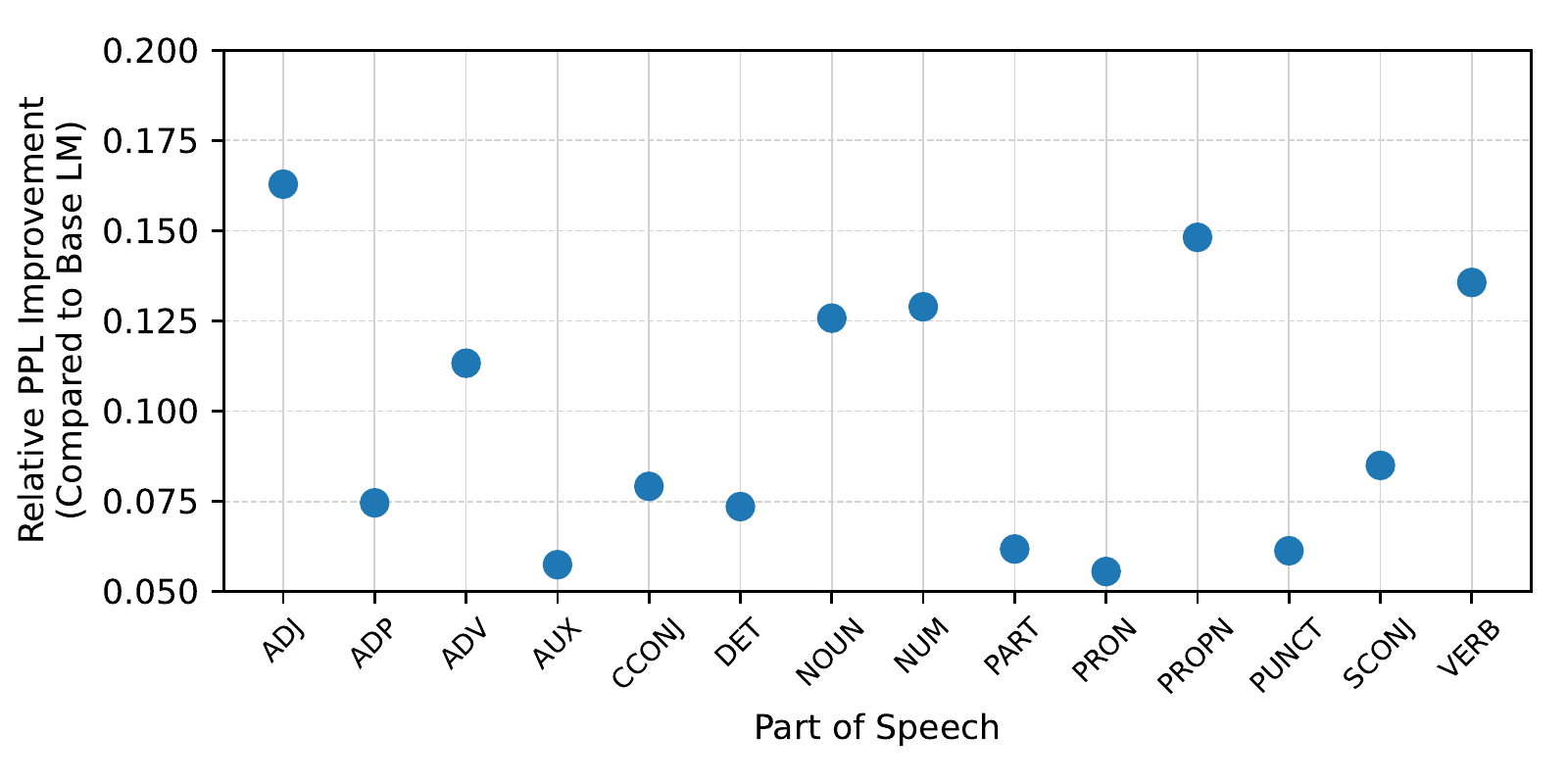}
\caption{Perplexity of the base language model grouped by part-of-speech (top), and relative improvement of the $k$NN-LM (bottom).}
\label{fig:analysis_pos_wiki_knnlm}
\end{figure}

The results are included in Figure \ref{fig:analysis_pos_wiki_knnlm}. We find that $k$NN-LM is most helpful for syntactic categories where the base language model most struggles, e.g. the original perplexity for adjectives (ADJ) is 105.37 and the $k$NN-LM improves perplexity by 16.3\% for this category. The five other categories that had worst perplexity (ADV, NOUN, NUM, PROPN, VERB) are also where $k$NN-LM works best.

This analysis serves as a useful sanity check. The syntactic categories are often associated with factual knowledge tied to entity relations, but no single category dominates performance. Also, there is some benefit for every category, so it is not clear that any should be avoided.

\begin{figure}[!ht]
\centering
\includegraphics[width=\linewidth]{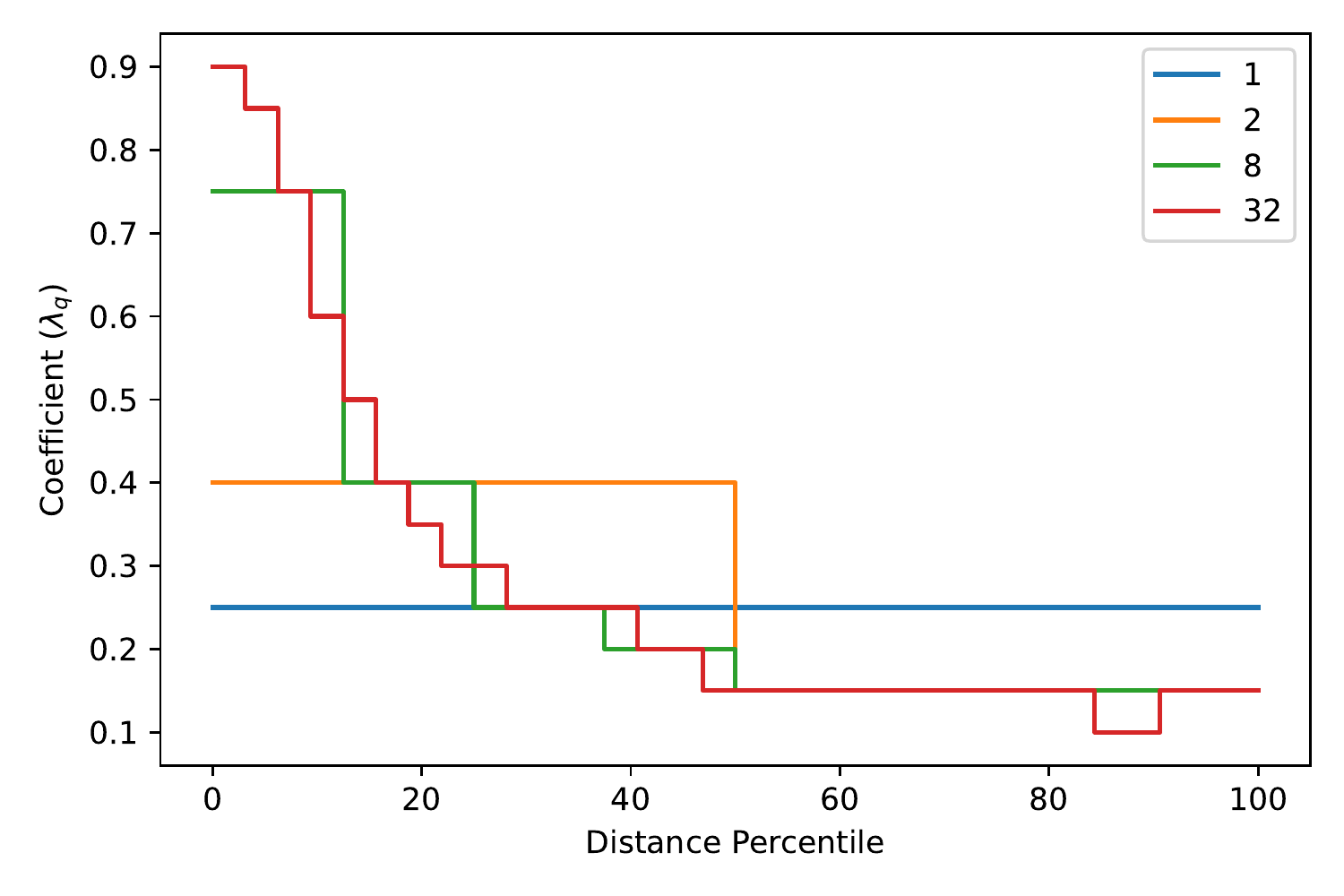}
\caption{Coefficient assignments ($\lambda_q$) after tuning on the Wikitext-103 validation set for different numbers of buckets, $b \in \{1, 2, 8, 32\}$.}
\label{fig:bucket_coeff}
\end{figure}

\section{A New Formulation for $k$NN-LM}

\label{sec:adaptive_coeff}

In the previous section, we analysed when $k$NN-LM is most helpful. We use this information to design a new formulation of $k$NN-LM that can exploit this behavior. The original $k$NN-LM uses the same interpolation coefficient ($\lambda$) for every example, which may not be desirable. As our analysis reveals, we can predict when the $k$NN-LM is most beneficial, which naturally leads us to a new formulation with an \emph{adaptive} $\lambda$:
\begin{align*} 
    P^{'}(w_t | .) &= \lambda_q P_{KNN}(w_t | .) + (1-\lambda_q) P(w_t | . )
\end{align*}
where $\lambda_q$ is a function of both the query and its retrieved documents rather than constant for all queries. This is highly similar to the formulation in \citet{He2021EfficientNN}, except theirs ignores retrieved items when deciding the coefficient.

Using the same $\lambda$ for all examples is limiting and does not leverage retrieval well if neighboring keys are clearly relevant (like shown in Figure \ref{fig:adaptive_coeff}). Of course, the critical decision here is how to map semantic similarity to an appropriate value for the coefficient.
We find it convenient and effective to use a piecewise function based on semantic similarity, following the bucketing described in \S\ref{sec:analysis_vecdist}.
We use the validation data for tuning, sorting by semantic similarity with the topic retrieved item then dividing all the queries into $b$ equally sized buckets. For each bucket we perform the same hyperparameter search over coefficients as in $k$NN-LM.\footnote{See \citealt{khandelwal20generalization} Figure 5.} 

Example coefficient assignments for different numbers of buckets ($b$) are shown in Figure~\ref{fig:bucket_coeff}.

\section{Experiments and Results}

To measure the importance of retrieval quality in the $k$NN-LM, we evaluate our approach (\S\ref{sec:adaptive_coeff}) on two English language modeling datasets. The first is the Wikitext-103 corpus \cite{merity2016pointer} used by \citet{khandelwal20generalization}. The second is PG-19 \cite{rae2020compressive}, which we include because it consists of books and is thematically distinct from the encyclopedic documents in Wikitext-103.

\subsection{Experimental Setup and Pre-processing}

\paragraph{Wikitext-103} The data is split 103M/217K/245K tokens for training, validation, and test. We use the pretrained model from \citet{khandelwal20generalization}, and associated 267K word-level vocab.

\paragraph{PG-19} To understand when adapting the coefficient to retrieval quality is desirable compared with a static coefficient, we include PG-19 in our experiments. PG-19 consists of books and is thematically distinct form the encyclopedic douments in the Wikitext-103 data. We sample 2,000 books from the training corpus, which gives approximately 150M tokens and is close in size to Wikitext-103. We use the standard validation split (50 books) and test split (100 books). We use word-level tokenization with a 300K vocabulary derived from our constructed training split. We train our own model using the same architecture and hyperparameters from \citet{khandelwal20generalization}.

\paragraph{Baselines} We choose these baselines to isolate the effect of retrieval quality on the performance of the $k$NN-LM: the self-attentive adaptive input representation from \citet{baevski2018adaptive} as the base model, the original $k$NN-LM \cite{khandelwal20generalization}, and the continuous cache model \cite{Grave2017ImprovingNL} which retrieves from both the datastore and local context. As described in \S\ref{sec:knnlm_approach}, the datastore is built by encoding a large text corpus, in this case the training set. Although we use approximate neighbors, we compute the next word probability with \textit{exact} distance as this substantially boosts performance \cite{khandelwal20generalization}.\footnote{\citet{He2021EfficientNN} and \citet{Alon2022NeuroSymbolicLM} present efficient extensions of $k$NN-LM. We exclude these as baselines since they're designed with approximate vector distance in mind.}

\begin{table}[t!]
\setlength\tabcolsep{4pt}
\begin{center}
\begin{tabular}{ r | c | c | c }
\toprule
$b$ & Dev$_{0}$ & Dev$_{1}$ & Dev \\
\midrule
1 & 17.091 & 14.989 & 16.091 \\
2 & 16.909 & 14.854 & 15.933 \\
4 & 16.763 & 14.767 & 15.815 \\
8 & 16.665 & 14.727 & 15.743 \\
16 & 16.637 & 14.722 & 15.727 \\
32 & 16.629 & 14.722 & 15.721 \\
64 & 16.622 & 14.724 & 15.719 \\
128 & 16.619 & 14.724 & 15.715 \\
\bottomrule
\end{tabular}
\end{center}
\caption{Validation perplexity on Wikitext-103. Used for hyperparameter tuning.}
\label{tab:tune}
\end{table}

\subsection{Tuning $k$NN-LM Hyperparameters}

For the original formulation of $k$NN-LM there are two hyperparameters to tune: the number of items to retrieve ($k$) and the interpolation coefficient ($\lambda$). These are tuned on the validation set. We introduce an important hyperparameter for the number of buckets to use ($b$) and tune a new interpolation coefficient ($\lambda_q$) separately for each bucket. Since each bucket is assigned its own coefficient, the total number of hyperparameters grows with the number of buckets. Even so, our approach has about the same speed as the original $k$NN-LM both for parameter tuning and during inference. We make hyperparameter tuning efficient by caching expensive computation (see \S\ref{sec:tuning} for more details). At test time, selecting the coefficient is an $O(1)$ lookup based on the semantic similarity of the top neighbor.

To select the number of buckets ($b$), we use the first half of the validation data (Dev$_0$) to define partition boundaries, and find the best performing interpolation coefficient for each partition separately. Then we measure perplexity on the second half of the validation data (Dev$_1$) using those partition boundaries and coefficients. The choice of $b$ that gives the best perplexity on Dev$_1$ is the one we ultimately use. With $b$ chosen, we then re-compute the partition boundaries and corresponding coefficients using the full validation data (Dev), which is used to evaluate against the test data.

An example of tuning for $b$ on Wikitext-103 is shown in Table \ref{tab:tune}. Increasing $b$ always leads to better perplexity on Dev$_0$, albeit with diminishing returns. Since the partition boundaries and coefficients are chosen using Dev$_0$, it is not guaranteed that increasing $b$ improves perplexity on the held-out data (Dev$_1$). Although, tuning the partition boundaries and coefficients on the validation data does not guarantee improvement on the test data, in our experiments we find our adaptive coefficient is always as effective as the original static one.

\subsubsection{Computational Cost of Tuning}

\label{sec:tuning}

Our approach is nearly the same speed as the original $k$NN-LM both at test time and for hyperparameter tuning. This is the case even though our hyperparameter count scales with $b$ and is more than an order of magnitude more than what is used for the $k$NN-LM. We accomplish this by effectively caching query vectors, retrieved items, and associated vector distances. The initial time to compute these values takes hours and is the same as with $k$NN-LM, but after computed it takes less than 5 minutes to perform the hyperparameter search for the adaptive coefficient on the Wikitext-103 data.\footnote{All experiments are run on a single Titan X GPU with 256GB CPU memory.} Our implementation with caching is available here: \href{https://github.com/iesl/knnlm-retrieval-quality}{github.com/iesl/knnlm-retrieval-quality}.

\begin{table}[t!]
\setlength\tabcolsep{4pt}
\begin{center}
\begin{tabular}{ l | c | c | c | c | c }
\toprule
 & $\lambda$ & $b$ & $k$ & Dev & Test \\
\midrule
Base LM & -    & - & -     &  17.96 & 18.65 \\
$k$NN-LM  & 0.25 & 1 &  1024 &  16.06 & 16.12 \\
+CCache & 0.25 & 1 &  1024 &  15.81 & 15.79 \\
\midrule
Ours (TFIDF) & $\lambda_q$ & 32 & 1024 & 15.76  & 15.54 \\
Ours         & $\lambda_q$ & 32 & 1024 & 15.72  & \textbf{15.50} \\
\bottomrule
\end{tabular}
\end{center}
\caption{Test and validation perplexity on Wikitext-103. This is our main result and demonstrates that our new formulation with adaptive coefficient ($\lambda_q$) substantially improves over $k$NN-LM.}
\label{tab:results_ppl}
\end{table}

\subsection{Perplexity on WikiText-103}

Table \ref{tab:results_ppl} reports the perplexity from our approach and various baselines on the Wikitext-103 validation and test sets. Our approach scores 15.50 perplexity on the test set. This is a 16.9\% improvement over the base language model and a 3.8\% improvement over the original $k$NN-LM formulation.

For the number of buckets ($b$) we found 32 to work best (see Table \ref{tab:tune}), and the set of coefficients are the same as shown in Figure~\ref{fig:bucket_coeff}. Our search space includes $b \in \{1, 2, 4, 8, 16, 32, 64, 128\}$ and $\lambda_q \in \{0.05, 0.1, 0.15, \dots , 0.9, 0.95\}$.

\citet{khandelwal20generalization} find that retrieving from recent history using the continuous cache model (CCache; \citealt{Grave2017ImprovingNL}) is complementary to retrieving from the datastore, improving perplexity when combined with $k$NN-LM. This type of caching is out of scope of this paper, and our approach already outperforms their combined model.

\subsection{Perplexity on PG-19}

To further understand how lexical overlap influences $k$NN-LM performance we evaluate using the PG-19 dataset. Compared to Wikipedia, text across books has much less repetition, so text retrieved from the datastore is less likely to overlap with $n$-grams in the evaluation data.

We train our own model using the same architecture and hyperparams for Wikitext-103, and report perplexity in Table \ref{tab:overlap_ppl}. We found $b =32$ works best. Despite the challenging properties of the book data, $k$NN-LM is still effective. Our re-formulation is marginally beneficial here.

\subsection{Filtering $n$-grams from the Datastore}

Thus far, our analysis indicates that lexical overlap is important for strong $k$NN-LM performance. To test this directly for our adaptive coefficient, we follow the procedure described in \S\ref{sec:analysis_bow} to rebuild the datastore but remove from the index large $n$-grams ($n \geq 8$) and their surrounding tokens that also appear in the evaluation data.

The results for this experiment on both Wikitext-103 and PG-19 are shown in Table \ref{tab:overlap_ppl}. Most of $k$NN-LM's improvements on Wikitext-103 come from retrieving contexts with overlapping $n$-grams,\footnote{As others have previously noted, Wikitext-103 contains considerable amounts of duplicate text \cite{McCoy2021HowMD}. Deduplicating the training data can be helpful for language modeling \cite{lee-etal-2022-deduplicating,dedup2022kandpal}, and sometimes other tasks \cite{schofield-etal-2017-quantifying}, but we completely remove text that overlaps with the evaluation data.} which could motivate simpler and faster retrieval functions. On the other hand, the cases in which $n$-gram overlap does not play a major role require further investigation. 

\section{Discussion}

In previous sections we use observations of $k$NN-LM to motivate our new approach that adapts the interpolation coefficient to retrieval quality. Here we analyze results with our new method to see how they compare with baselines and deepen our understand of retrieval-enhanced language modeling.

\begin{table}[t!]
\setlength\tabcolsep{4pt}
\begin{center}
\begin{tabular}{ l | c | r | r | c }
\toprule
 & $\lambda$ & $b$ & $k$ & Dev \\
\midrule
Dense & 0.25        &  1 & 1024 & 16.06 \\
Dense & $\lambda_q$ & 32 & 1024 & 15.72 \\
TFIDF & $\lambda_q$ & 32 & 1024 & 15.76 \\
\midrule
Dense & 0.05  & 1 & 1   & 17.10 \\
Dense & 0.15  & 1 & 8   & 16.66 \\
Dense & 0.25  & 1 & 64  & 16.31 \\
\midrule
Dense & $\lambda_q$ & 16  & 1   & 16.63     \\
Dense & $\lambda_q$ & 128 & 8   & 16.19    \\
Dense & $\lambda_q$ & 16  & 64  & 15.90   \\
\midrule
TFIDF & $\lambda_q$ & 32  & 1   & 16.38     \\
TFIDF & $\lambda_q$ & 64  & 8   & 16.06    \\
TFIDF & $\lambda_q$ & 16  & 64  & 15.87   \\
\bottomrule
\end{tabular}
\end{center}
\caption{Validation perplexity on Wikitext-103 used for ablation analysis. The $k$NN-LM uses a single static value for the interpolation coefficient ($\lambda$), our method uses an adaptive coefficient ($\lambda_q$). This table includes our approach when using the semantic similarity (Dense) or bag-of-words representation (TFIDF). Based on how many items are retrieved ($k$), our approach works best with a different amount of buckets ($b$).}
\label{tab:compare_tfidf}
\end{table}

\subsection{Can we adapt to lexical similarity?}

The original $k$NN-LM has similar performance when its results are stratified by either semantic or lexical similarity (\S\ref{sec:analysis_vecdist}), but in our new formulation we adaptive the coefficient only according to semantic similarity. What if we use lexical similarity instead? We explore this possible alternative and report the results for Wikitext-103 in Table \ref{tab:compare_tfidf}.

In general, we find that both semantic and lexical similarity\footnote{To measure lexical similarity we use TFIDF vectors.} yield similar results when used to bucket queries. For the best setting, when $k=1024$, the learned vectors work better, reflecting recent findings that dense vectors outperform sparse representations for various retrieval-related tasks \cite{lee-etal-2019-latent,gao-etal-2021-coil}. Hence, throughout this paper we adapt the coefficient using semantic similarity and $k=1024$ unless otherwise specified. Interestingly, for lower values of $k$ the bag-of-words representation has an edge over semantic similarity. Perhaps this suggests lexical similarity is more precise, and if retrieving many items is costly, then adapting the coefficient according to lexical similarity might be particularly helpful. 

\begin{figure}[!ht]
\centering
\includegraphics[width=\linewidth]{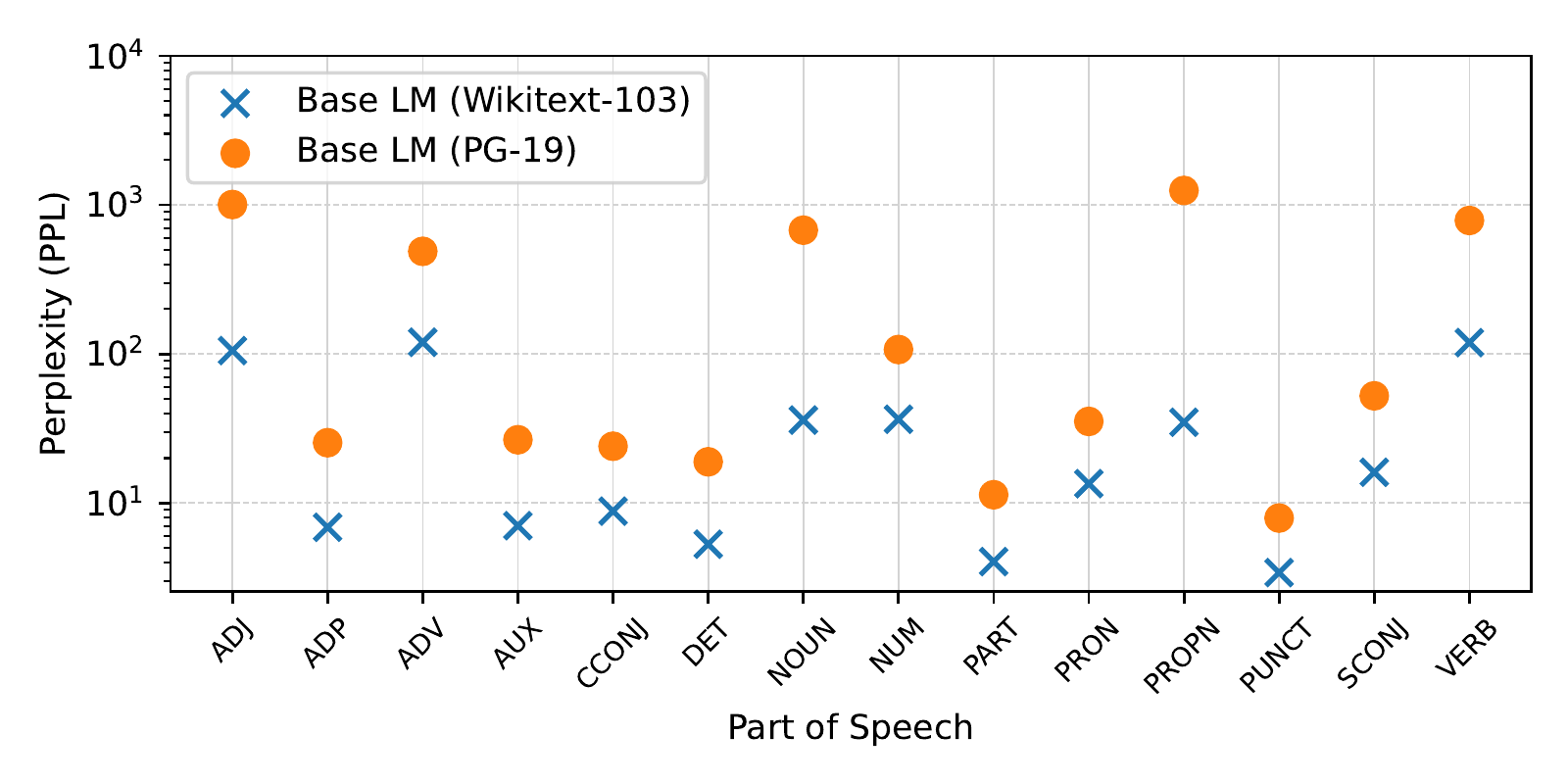}
\includegraphics[width=\linewidth]{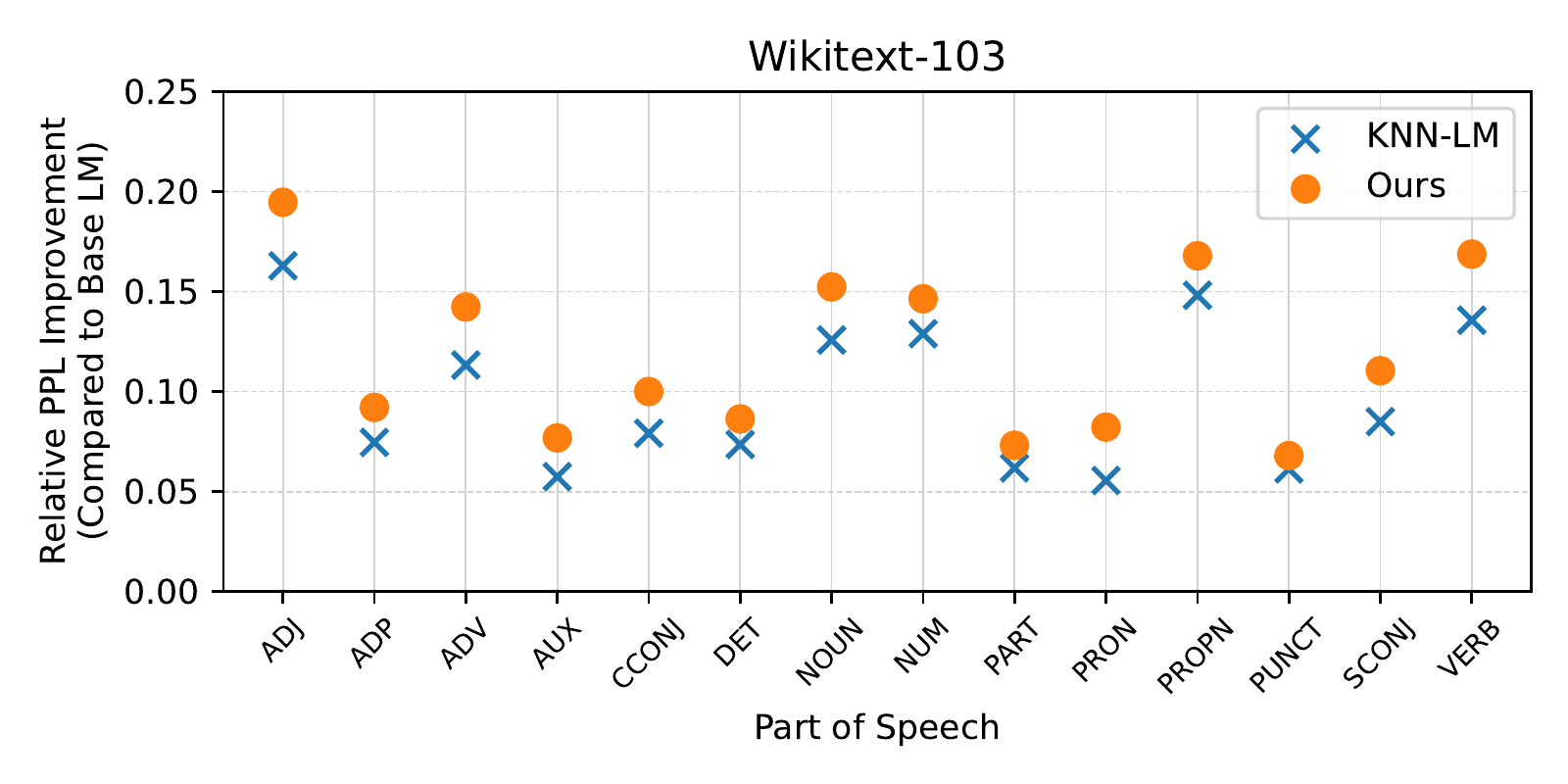}
\includegraphics[width=\linewidth]{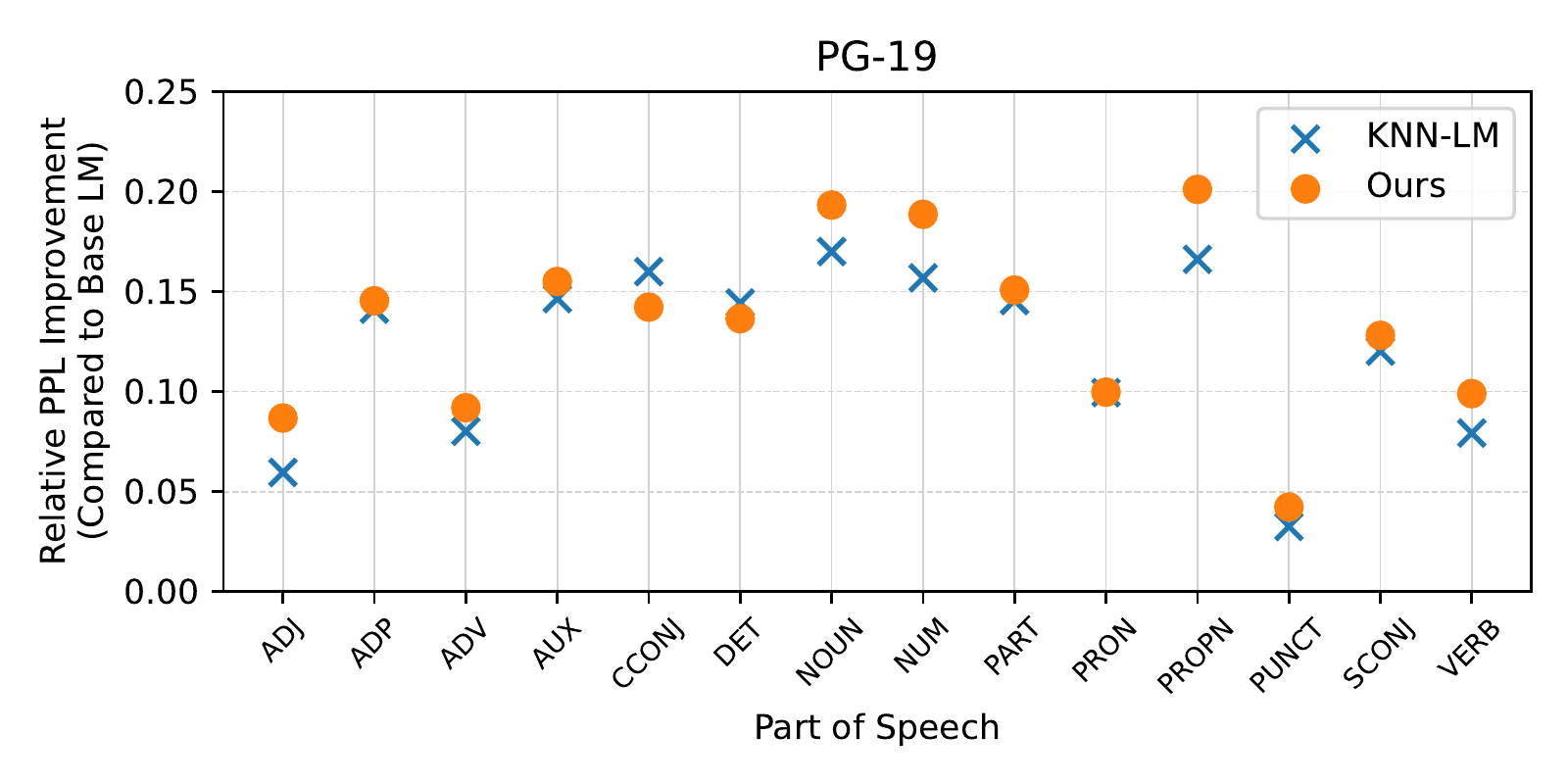}
\caption{Perplexity of the base language model (top), grouped by part-of-speech. Relative perplexity improvement by $k$NN-LM approches on Wikitext-103 (center) and PG-19 (bottom). The lines corresponding $k$NN-LM match Figure \ref{fig:analysis_pos_wiki_knnlm} --- they are included here to emphasize the difference to our new formulation.}
\label{fig:pos_ppl_combined}
\end{figure}

\begin{table*}
\setlength\tabcolsep{4pt}
{\small
\begin{center}
\begin{tabular}{l l l}
\toprule[1.5pt]
Book & Context & $r$ \\
\midrule[1pt] 
{\small \textit{The Unbearable Bassington}, Saki (1912)} &
My dear Francesca , he said soothingly , laying his hand \textbf{affectionately} &
{\small $q$}
\\
\midrule[0.1pt]
{\small \textit{FLORA}, A.L.O.E. (1860)} &
My dear madam , said Mr. Ward earnestly , laying his hand \textit{on} &
{\small $1$} 
\\
{\small \textit{Peter}, Smith (1908)} &
this young man 's uncle , said Peter , laying his hand \textbf{affectionately} &
{\small $11$}
\\
\midrule[1pt] 
{\small \textit{Life of Napoleon Bonaparte}, Sloane (1896)} &
during the worst periods of terror , were thronged from pit to \textbf{gallery} &
{\small $q$}
\\
\midrule[0.1pt]
{\small \textit{Sketches of Reforms\textemdash}, Stanton (1849)} &
For weeks , that theater was crowded from pit to \textit{dome} &
{\small $1$}
\\
{\small \textit{Farquharson of Glune}, Bateman (1908)} &
The storm of feeling swept alike from stall to \textbf{gallery} &
{\small $6$}
\\
\midrule[1pt] 
{\small \textit{Walking}, Thoreau (1851)} &
like a dream of the Middle Ages . I floated down its historic \textbf{stream} &
{\small $q$}
\\
\midrule[0.1pt]
{\small \textit{The Automobilist Abroad}, Mansfield (1907)} &
 France is a pleasure , a voyage up a picturesque and historic \textit{French} &
{\small $1$}
\\
{\small \textit{Canadian Notabilities}, Dent (1880)} &
two small sailing craft slowly making their way up the majestic \textbf{stream} &
{\small $42$}
\\
\bottomrule[1.5pt]
\end{tabular}
\end{center}
}
\caption{Examples from PG-19 where relevant contexts are found even with large $n$-grams removed from the datastore. There can be overlap in small $n$-grams (top), local structure (center), or semantics (bottom). The contexts are shown with their corresponding book. Rank ($r$) is shown except for queries ($q$). Values are bolded or italicized.}
\label{tab:data_examples}
\end{table*}

\subsection{Do syntactic trends hold across domains?}

We repeat the syntactic analysis from \S\ref{sec:syntactic_analysis} using our adaptive coefficient and include PG-19 as an additional dataset.\footnote{We only include the first 500K tokens from PG-19 validation data, as this is already more than twice the size of Wikitext-103 validation data.} The corresponding plots are shown in Figure \ref{fig:pos_ppl_combined}.

In both domains, the base model has a similar pattern of perplexity for part-of-speech tags, but there are some differences when comparing $k$NN-LM across domains. For instance, $k$NN-LM is especially helpful for adjectives in wikipedia text, but much less so for the book data. It's satisfying to see our new formulation of the $k$NN-LM has a similar impact in many cases for both domains, e.g. improving performance on adjectives nearly 5\% despite the aforementioned differences. Also, our formulation and $k$NN-LM provide consistent benefits even in the relatively more challenging book domain. Besides being potentially stylistically and syntactically distinct, we imagine encyclopedic text has more repetition than book data, which would likely influence the amount of lexical overlap between the train and evaluation data. We explore the effect of deliberately limiting lexical overlap in the next subsection, providing insights for the different cases when retrieval is helpful.

\subsection{What use is the restricted datastore?}

As we established in \S\ref{sec:analysis_bow}, the lexical overlap between a query and a retrieved context is a reasonable proxy for relevance. In Table \ref{tab:overlap_ppl}, we report the perplexity of our adaptive coefficient when ignoring large $n$-grams that overlap with the evaluation data when building the index, yielding a restricted less effective datastore. With these highly \textit{relevant} contexts removed, we observe that the $k$NN-LM shows substantially worse test perplexity on Wikitext-103, 18.05 instead of 16.12. PG-19 exhibits different behavior, and the change in perplexity is minimal. This suggests that $k$NN-LM can be helpful even when there are not large overlapping $n$-grams between the datastore and evaluation corpus --- such cases occur frequently in PG-19, and we visualize examples in Table \ref{tab:data_examples}.

With the restricted datastore, the benefit from adapting the coefficient is substantially diminished for Wikitext-103, but less so for PG-19. This suggests the partitions capture qualities besides lexical similarity. Alternatively, it could be that short $n$-grams are helpful in Wikitext-103, despite \citet{khandelwal20generalization} reporting that interpolating the base language model with an $n$-gram model was not very effective.

It is worth noting that even when contexts with high lexical overlap are removed from the datstore, adapting the coefficient is robust and provides performance at least on par with $k$NN-LM in the same setting. While $k$NN-LM is weakened here, it does improve over the base language model. In future work, it could prove fruitful to explore alternate strategies besides semantic or lexical similarity.

\section{Related Work}

We extend the $k$NN-LM by adapting the interpolation coefficient to retrieval quality (measured by semantic similarity).
AdaptRet \cite{He2021EfficientNN} models the interpolation coefficient as a function of the query. This is convenient, since one can skip retrieval if the coefficient is below a threshold, although requires training a separate adaptor network. Crucially, their coefficient predictions are based solely on query features, and does not take into account whether retrieval is successful. Our approach incorporates the quality of retrieval, and improves language modeling results. It is simple and effective, and only needs lightweight hyperparameter tuning without any additional training.

RetoMaton \cite{Alon2022NeuroSymbolicLM} provides an alternative means to bypass retrieval. They build a graph over the datastore, and at each time step they either retrieve like the original $k$NN-LM or re-use the previously retrieved neighbors to traverse the graph. This is more efficient than AdaptRet, providing better results at lower cost. Both AdaptRet and RetoMaton are designed with efficiency in mind. They rely on approximate distance using product quantization and perform about as well as the exact distance version of the $k$NN-LM. We improve upon $k$NN-LM by about 4\% perplexity.

There are many recent works that use retrieval components for language tasks besides language modeling, such as question answering \citep{Godbole2019MultistepEI,Guu2020REALMRL, kassner-schutze-2020-bert}, dialogue generation \citep{Fan2021AugmentingTW}, conversational search \cite{Hashemi2020GuidedTL}, semantic parsing \citep{Gupta2021RETRONLURA}, data augmentation \cite{Du2021SelftrainingIP}, and machine translation \cite{khandelwal2021nearest,zheng-etal-2021-adaptive,chunkMT2022}. 

There are alternatives to $k$NN-LM that incorporate document structure \cite{Xu2021CapturingSL}, but their experimental setup is not comparable with ours. In our baselines we only consider models matching the original $k$NN-LM backbone, although alternative architectures show promise for retrieval-enhanced language modeling \cite{Yogatama2021AdaptiveSL,meng2022gnnlm,zhong2022training}. Scaling the datastore \cite{Borgeaud2021ImprovingLM} or model size \cite{Shoeybi2019MegatronLMTM} have shown to effectively improve language modeling. Alternatively, text generation may be improved through more advanced ranking \cite{min-etal-2021-joint} or decoding \cite{rankgen22} algorithms.

Researchers have explored fundamental extensions to $k$NN that are agnostic to language data. \citet{locallyadaptive1993} spatially partition the datastore, adapting the value of $k$ for each region. Keeping $k$ fixed, \citet{discadapt1995knn} instead adapt the shape of the neighborhood based on local information. 

\section{Conclusion}

In this paper, we have  proposed a novel and effective re-formulation of the $k$NN-LM. Our approach adapts the interpolation coefficient to the quality of retrieved documents measured by semantic similarity. We motivate our approach through extensive analysis, which also provides insights on the types of tokens and contexts $k$NN-LM is most helpful for. Importantly, we empirically demonstrate the effectiveness of our approach through experiments on two domains, Wikitext-103 (encyclopedic text) and PG-19 (book data), and outperform the original $k$NN-LM by 4\% test perplexity on the Wikitext-103 language modeling corpus.

\section*{Limitations}

The $k$NN-LM leverages a datastore, and when populated with text relevant for the task domain, can be used to improve language modeling performance. The benefits of this procedure are data dependent and domain-specific, and the same applies to the adaptive coefficient technique that we introduce.

The adaptive coefficient requires many more tunable hyperparameters. To address this, we release an optimized codebase to perform this hyperparameter search in neglible time compared with the original $k$NN-LM.

\section*{Ethical Concerns and Impact}

Even when used with the best intentions language models can produce malicious or harmful text, and guards are typically used to account for inherent bias or undesirable output. In our case, we do not generate text and simply use the model to evaluate perplexity on existing data, so effectiveness of safety guards and their limitations is not a relevant concern in this work.

\section*{Acknowledgements}

We are grateful to Fernando Diaz, Urvashi Khandelwal, Kalpesh Krishna, Simeng Sun, the UMass NLP group and IESL for several useful discussions during the course of the project. This work was supported in part by the Center for Intelligent Information Retrieval and the Center for Data Science; in part by the IBM Research AI through the AI Horizons Network; in part by the Chan Zuckerberg Initiative under the project Scientific Knowledge Base Construction; in part by the National Science Foundation (NSF) grant numbers  IIS-1922090, IIS-1955567, IIS-1763618, and IIS-2106391; in part by the Defense Advanced Research Projects Agency (DARPA) via Contract No. FA8750-17-C-0106 under Subaward No. 89341790 from the University of Southern California; and in part by the Office of Naval Research (ONR) via Contract No. N660011924032 under Subaward No. 123875727 from the University of Southern California. Any opinions, findings and conclusions or recommendations expressed in this material are of the authors and do not necessarily reflect those of the sponsor.

\bibliography{anthology,acl}
\bibliographystyle{acl_natbib}

\end{document}